\documentclass[pre,twocolumn,twoside,showpacs,superscriptaddress,floatfix]{revtex4-1}
\usepackage{graphics,amssymb,amsmath,epsf,color,hyperref}
\epsfclipon
\usepackage{epsfig,bm,epsf,graphics}
\usepackage{graphicx}
\usepackage{dcolumn}
\usepackage{bm}
\usepackage{braket}
\usepackage{color}
\usepackage{soul}
\usepackage{xcolor}
\usepackage{floatrow}
\usepackage{hyperref}

\hypersetup{colorlinks, citecolor=blue, linkcolor=blue}

\newcommand{\revi}{\textcolor{black}}
\newcommand{\revf}{\textcolor{black}}

\usepackage{graphicx}
\newcommand{\indep}{\rotatebox[origin=c]{90}{$\models$}}

\begin{document}

\title{Scale-invariant representation of machine learning}

\author{Sungyeop Lee}
\email[Corresponding author: ]{dtd2001@snu.ac.kr}
\affiliation{Department of Physics and Astronomy, Seoul National University, Seoul 08826, Korea}

\author{Junghyo Jo}
\email[Corresponding author: ]{jojunghyo@snu.ac.kr}
\affiliation{Department of Physics Education and Center for Theoretical Physics and Artificial Intelligence Institute, Seoul National University, Seoul 08826, Korea}
\affiliation{School of Computational Sciences, Korea Institute for Advanced Study, Seoul 02455, Korea}

\date{\today}

\begin{abstract}
The success of machine learning has resulted from its structured representation of data.
Similar data have close \revi{internal} representations as compressed codes for classification or emerged labels for clustering.
We observe that the frequency of internal \revi{codes or labels} follows power laws in both supervised and unsupervised learning models.
This scale-invariant distribution implies that machine learning largely compresses frequent typical data, and simultaneously, differentiates many atypical data as outliers.
In this study, we derive the process by which these power laws can naturally arise in machine learning.
In terms of information theory, the scale-invariant representation corresponds to a maximally uncertain data grouping among possible representations that guarantee a given learning accuracy.
\end{abstract}

 
\maketitle


\section{Introduction}

The remarkable performance of machine learning~\cite{lecun2015, ghahramani2015,carleo2019} is due to the internal representation, denoted by $z$, of features extracted from data by neural network  models.
Here, $z$ functions as effective representations to discriminate images, speech, time series for pattern recognition~\cite{lecun1995}, classical and quantum phases in matters and active matters~\cite{van2017, carrasquilla2017, rem2019, cichos2020}, chemical structures for drug discovery~\cite{sanchez2018,noe2019}, time arrows of non-equilibrium dynamics~\cite{seif2021}, etc. 
Therefore, understanding the mechanism behind the effective feature distillation of neural network models is a fundamental problem in machine learning.  

The information bottleneck theory interprets machine learning as information compression and transmission as described by communication theory~\cite{tishby2000}.
Neural networks encode internal representations $z$ that maximally compress irrelevant information in input data $x$ to restore desired outputs $y$, called labels.
Given data without labels, autoencoders use input itself as output $y=x$~\cite{kramer1991}.
Then, the self-supervised learning can perform the dimensional reduction of data by providing a compressed representation $z$ that can faithfully reproduce $x$. 
In particular, when the transformation of $x \rightarrow z$ is a linear mapping, the machine corresponds to principal component analysis~\cite{bourlard1988, baldi1989}. In addition to the self-supervised learning, unsupervised learning such as restricted Boltzmann machines (RBMs) and deep belief networks have also been used for dimensional reduction of $x$~\cite{hinton2006}. In unsupervised learning, the internal representation $z$ can be interpreted as emergent labels for each $x$. 

The representations $z$ sometimes reflect features themselves, edges detected in image recognition~\cite{lindsey2019}. On the other hand, $z$ can be considered as a dummy code, the frequency of which only matters when no prior knowledge of $x$ is provided.
\revf{Let us focus on compressing structures of neural networks in which internal layers have lower dimensions as they are farther from input layers.}
Interestingly, \revf{given the compressing structures,} a recent work observed that the frequency of $z$ follows power laws in \revi{RBMs}~\cite{powerlaw,rule2020}, reminiscent of criticality in statistical mechanics~\cite{cubero2019}. Using the dummy code $z$ as emergent labels of $x$, the frequency of $z$ can be interpreted as a cluster size of $x$ labeled by $z$. Then, the power laws imply that the cluster size distribution is scale-invariant.
It is an interesting challenge to address how these power laws arise during the learning process devoid of any specific instruction for scale-invariance. 
\revi{Song {\it et al.} have derived that the power-law clustering is an entropy-maximized distribution at a certain compression level of $z$ in RBMs~\cite{powerlaw}.
In this study, we extend this idea using information theory, and show that the power-law scaling of the cluster size distribution emerges naturally not only in unsupervised learning, but also in supervised learning.
}

This paper is organized as follows. In Sec.~\ref{sec:experiment}, we show that various machine learning models have scale-invariant distributions for their internal representation. 
We then explain the emergence of this scale invariance using information theory in Sec.~\ref{sec:theory}. Finally, we summarize and discuss our findings in Sec.~\ref{sec:summary}.

\begin{figure*}[t]
\includegraphics[scale=0.25]{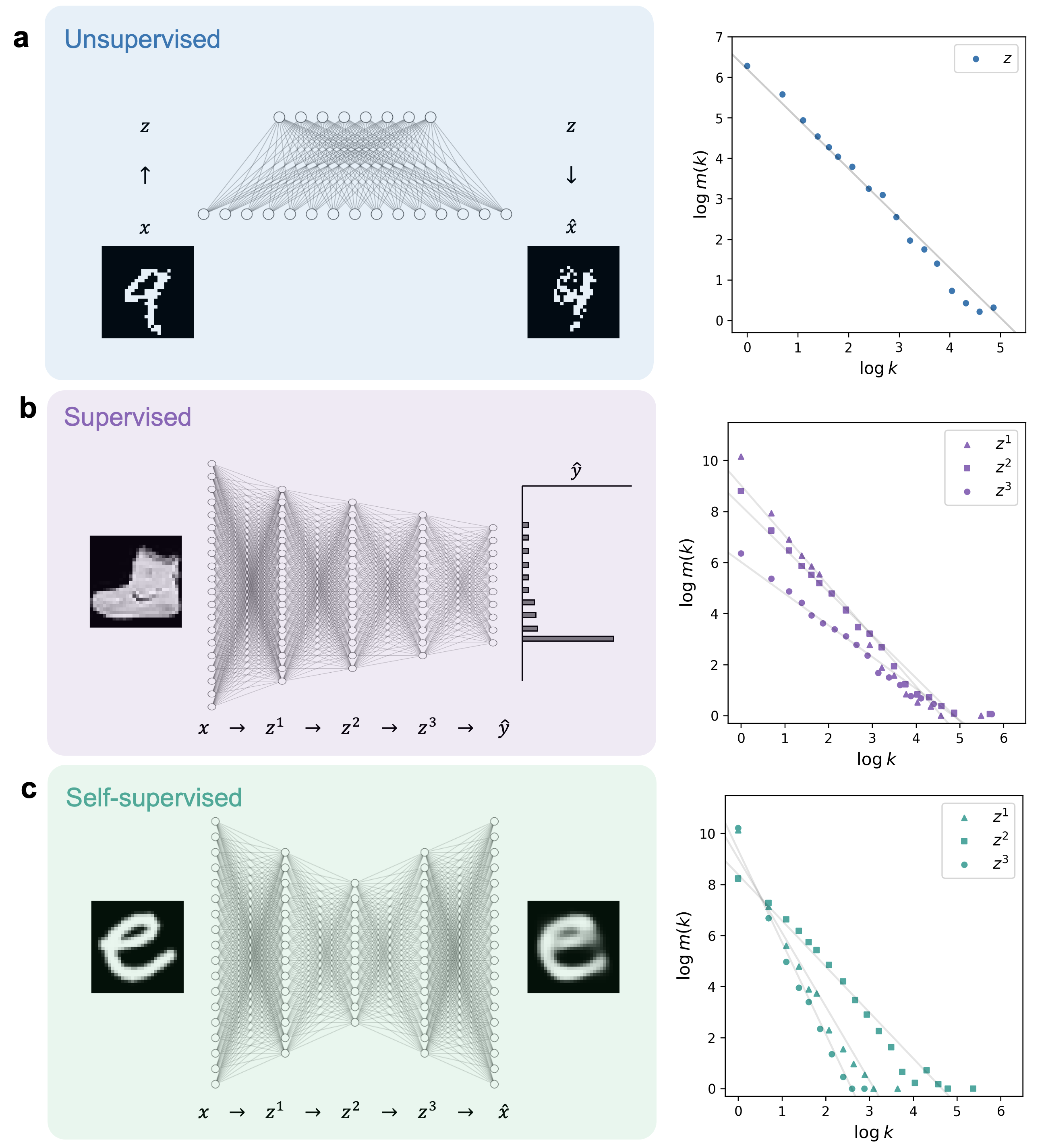}
\centering
\caption{\revi{(Color online) Scale-invariant internal representations of machine learning. (a) Unsupervised learning of restricted Boltzmann machine (RBM) with MNIST dataset. 
The network architecture of RBM models comprises visible and hidden units of $x$ and $z$. 
An image $x$ of MNIST was reconstructed to $\hat{x}$ through the hidden representation $z$.
(b) Supervised learning with image dataset of Fashion-MNIST. The architecture of a deep neural network consists of input $x$, hidden $z^\mu(\mu=1,2,3)$, and output $\hat{y}$. Note that the number of nodes is arbitrary for a schematic visualization.
(c) Self-supervised learning with EMNIST dataset. The model had a symmetric structure with the bottleneck layer $z^2$.
Right panels show corresponding log-log plots between the degeneracy $m(k)$ and frequency $k$ of internal representations: $z$ (circles) for RBM; and $z^1$ (triangles), $z^2$ (squares), and $z^3$ (circles) for supervised and self-supervised learning.
}
}
\label{fig1}
\end{figure*}

\section{Power laws in machine learning}
\label{sec:experiment}
\revi{We first reproduce that the frequency of $z$ follows power laws in RBMs (Fig.~\ref{fig1}a). Next, we observe that the scale-invariant internal representation is also found in supervised learning (Fig.~\ref{fig1}b, c).
Finally, we examine how the emergence of power laws can depend on learning processes and data preparation. 
To aid in reproducing our results, we have provided the complete source code and documentation on GitHub~\cite{Github}.
}

\begin{figure*}[t]
\includegraphics[scale=0.23]{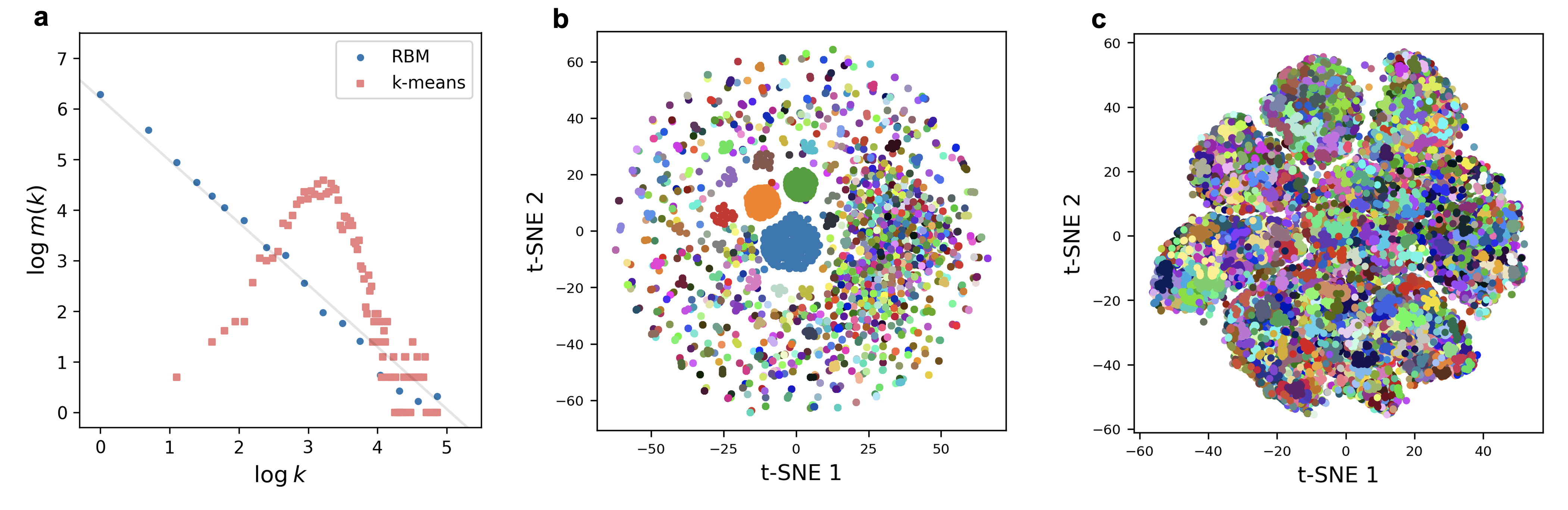}
\centering
\caption{\revi{(Color online) Power-law data clustering. (a) The size distribution of data clusters via internal representation $z$ of the restricted Boltzmann machine (RBM, blue circles) and k-means clustering (coral squares).
(b) Two-dimensional visualization of $z$ using t-distributed stochastic neighbor embedding (t-SNE) with different colors for different $z$.
(c) t-SNE plot of $x$ with different colors for different k-means clusters. We set k$=2048$ to be comparable with the total number ($\approx 2000$) of distinct states in the RBM.
}
}
\label{fig2}
\end{figure*}

\subsection{\revi{Unsupervised learning}}
The goal of unsupervised learning is to extract the inherent probability distribution $p(x)$ of data $x$. In contrast, supervised learning extracts the paired information between $x$ and label $y$. RBM is a representative neural network for unsupervised learning that is composed of an input layer for $x$ and a hidden layer for $z$~\cite{salakhutdinov2007, hinton2012}. RBMs have a special graph structure in which input nodes are not connected to other input nodes, and the same is true for hidden nodes (Fig.~\ref{fig1}a). This allows the factorization of a conditional probability $p(z|x)=\prod_i p(z_i|x)$ for $z=(z_1, z_2, \cdots, z_m)$ and $p(x|z)=\prod_j p(x_j|z)$ for $x=(x_1, x_2, \cdots, x_n)$.
The goal of RBM, that of matching the model distribution $p(x)$ into the data distribution $\hat{p}(x)$, is achieved by the contrastive divergence algorithm, a type of sampling method which uses the forward probability $p(z|x)$ and backward probability $p(x|z)$~\cite{CD}.

To perform unsupervised learning, we used the MNIST dataset~\cite{MNIST}, which consists of 60,000 training and 10,000 testing images of $28\times28$ pixels of 10 different hand-written digits (0-9).
After the RBM successfully generates original digit images, we counted the frequencies $k_z$ of specific $z$ allowing the mapping of different images of $x$ into the same $z$.
We observed that the neural networks exhibited a few frequent $z$ and many rare $z$, which is described by the degeneracy of the frequency $k$, $m(k) = \sum_z \delta \big(k- k_z\big)$.
It is of particular interest that the frequency degeneracy follows power laws of $m(k) \sim k^{-\beta -1}$ (Fig.~\ref{fig1}a), \revi{as reported in Song {\it et al.}~\cite{powerlaw}.}
The scale-invariant cluster size distribution was non-trivial given that the representative clustering method of k-means, based on the Euclidean distance between data, produced unipolar distribution with a characteristic cluster size (Fig.~\ref{fig2}a). 
\revi{The different cluster size distributions can be further visualized using two dimensional projection through t-distributed stochastic neighbor embedding (t-SNE) method (Fig.~\ref{fig2}b, c).
}
The scale-invariant distribution of RBMs may lead to a functional advantage in classifying data into a large cluster of frequent typical data, and many small clusters of atypical data as outliers.

\subsection{\revf{Supervised learning}}
\revi{The goal of supervised learning is to predict true labels $y$ from input data $x$.}
In communication theory ($y \rightarrow x \rightarrow z \rightarrow \hat{y}$), a message $y$ is transferred to a noisy code $x$, which is then mapped into a compressed code $z$. Finally, we decode $z$ to obtain $\hat{y}$. The transmission succeeds if the decoded message is consistent with the true message ($\hat{y}=y$).

As a concrete example, we consider the Fashion-MNIST dataset~\cite{FMNIST}, in which labels $y$ are assigned to ten fashion products such as sneakers and shirts. The labels are assigned to 70,000 (60,000 training and 10,000 test) $28 \times 28$ pixel images of $x$. Using a deep neural network, we transformed $x \rightarrow z^1 \rightarrow z^2 \rightarrow z^3 \rightarrow \hat{y}$ while reducing the dimension of the corresponding layers from $784$ to $70, 50, 35,$ and $10$ (Fig.~\ref{fig1}b).
Here, true labels $y$ are expressed as 10-dimensional one-hot vectors, the components of which have values between 0 and 1.
The network is trained to reduce the discrepancy between $y$ and $\hat{y}$.

Once the classification accuracy for the test data reached 87\%, we examined frequencies of internal representations of $z^1, z^2,$ and $z^3$.
Because these representations have continuous values in multi-layer perceptrons, we binarized them to count finite coarse-grained representations. 
We counted frequencies $k_z$ of \revi{discretized} $z$ allowing the mapping of different images of $x$ into the same $z$.
It is of particular interest that the frequency degeneracy followed power laws of $m(k) \sim k^{-\beta -1}$, where the exponent $\beta$ depends on the dimension of the $\mu$-th hidden layers $z^\mu$ (Fig.~\ref{fig1}b).
\revi{We confirmed that the existence of power laws was insensitive to the binarization threshold, and the power-law exponents did not depend on the initialization of learning.
Furthermore, our conclusion was robust on varying the architecture depth and width of neural networks, if they had narrowing structures for compressing.}
It is noteworthy that the scale invariance had never been instructed by the learning algorithm.

\subsection{\revi{Self-supervised learning}}
In real-world datasets, labels are not always available. For such datasets, if $x$ plays the role of label $y=x$, self-supervised learning models are referred to as {\it autoencoders}, which are useful for dimensional reduction~\cite{kramer1991}, denoising~\cite{vincent2008}, and generation~\cite{kingma2013, kingma2019}. In particular, the structure of $x \rightarrow z \rightarrow \hat{x}$ implies that the compressed representation $z$ is used to reconstruct the original $x$ (Fig.~\ref{fig1}c).
\revi{This study focuses on the compressing conditions where the dimension of $z$ is smaller than that of $x$.
Unlike the narrowing networks, widening networks map different $x$ into different $z$. Hence, the frequency of representation $z$ is trivial, with $k_z = 1$.
}

\revi{Autoencoders are closely related to RBMs.
}
The unfolded structure of the forward and backward processes of $x \leftrightarrow z$ in RBMs can be interpreted as information flows of  $x \rightarrow z \rightarrow \hat{x}$ in autoencoders where the weight parameter of the encoder part is the transpose of the weight parameter of the decoder part~\cite{AE_RBM}. 
\revi{However, it is noteworthy that $z$ is a stochastic and discrete variable in RBMs, whereas $z(x)$ is a deterministic and continuous function of $x$ in autoencoders. Nevertheless, if one adopts the sigmoid function as the activation function for autoencoders, $z(x)$ can be interpreted as the expectation value $\mathbb{E}[z]$ of $z$ in RBMs.}

We confirmed that \revi{binarized} $z$ of the autoencoders also showed the power-law scalings when we carried out all-to-all connected multilayer perceptron learning on the above Fashion-MNIST data (data not shown). Instead of repeating with the same dataset, here we considered another dataset, EMNIST, consisting of hand-written images of 10 digits (0-9) and 26 uppercase (A-Z) and lowercase letters (a-z)~\cite{EMNIST}. Among them, we trained 30,000 lowercase letters with deep structure of autoencoders $x \rightarrow z^1 \rightarrow z^2 \rightarrow z^3 \rightarrow \hat{x}$. Once the reconstructed image $\hat{x}$ faithfully reproduced the original image $x$, we counted the frequency of $z^\mu$ of \revi{each hidden} layer. The frequency degeneracy $m(k)$ again followed power laws \revi{in every hidden layer with different exponents} (Fig.~\ref{fig1}c). 

\begin{figure}[t]
\includegraphics[scale=0.2]{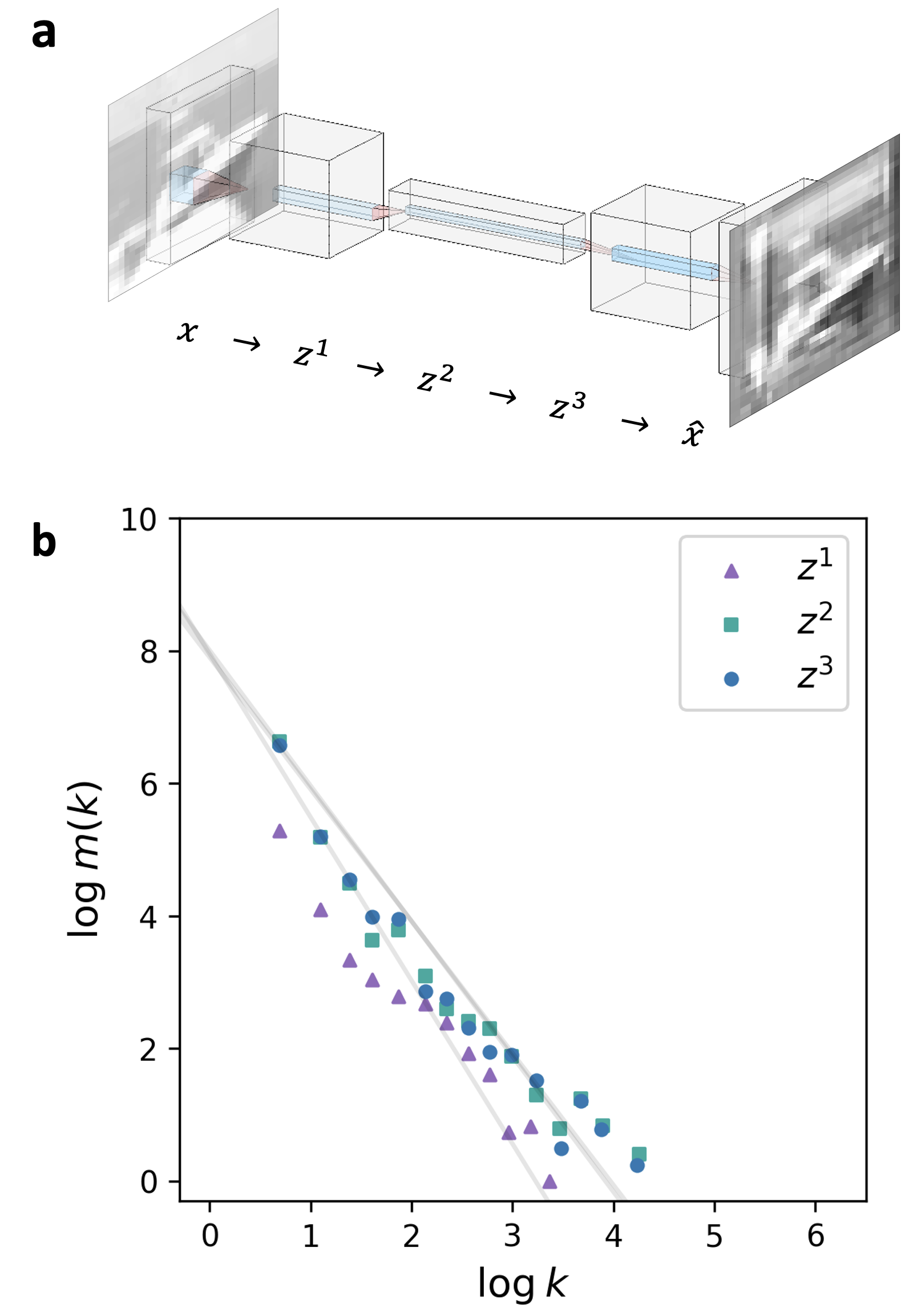}
\centering
\caption{\revi{(Color online) Internal representations of a convolutional neural network (CNN).
(a) Architecture of a CNN with three internal layers of $z^1, z^2,$ and $z^3$.
Image data of CIFAR-10 reconstructed through the self-supervised learning of the CNN.
(b) Frequency distributions of internal representations: $z^1$ (purple triangles), $z^2$ (green squares), and $z^3$ (blue circles).
}
}
\label{fig3}
\end{figure}

\revi{To examine the robustness of our findings, we replaced the sigmoid activiation function with a rectified linear unit (ReLU) function. Then, we confirmed that power laws remained evident with the ReLU activation (data not shown).
Next, we examined a convolutional neural network (CNN) that incorporated the information of proximal sites of $x$, which is known to show excellent performance for image recognition~\cite{lecun1998}.
We considered CIFAR-10, consisting of 50,000 training and 10,000 testing 32$\times$32 color images in 10 classes, such as airplane and automobile~\cite{CIFAR}. We then adopted convolutional autoencoders.
Once the reconstructed image $\hat{x}$ faithfully reproduced the original image $x$, we counted the frequency of $z$, and confirmed that the frequency degeneracy $m(k)$ again followed power laws (Fig.~\ref{fig3})}, \revf{although they were less prominent than the previous ones in Fig.~\ref{fig1}.
For the successful image reconstruction, the CNN is required to have high-dimensional internal representations with many filters.
On the other hand, the increased model complexity caused undersampling for a given data. 
The trade-off hindered a clearer confirmation of the existence of power laws in the CNN.
}

\begin{figure}[t]
\includegraphics[scale=0.23]{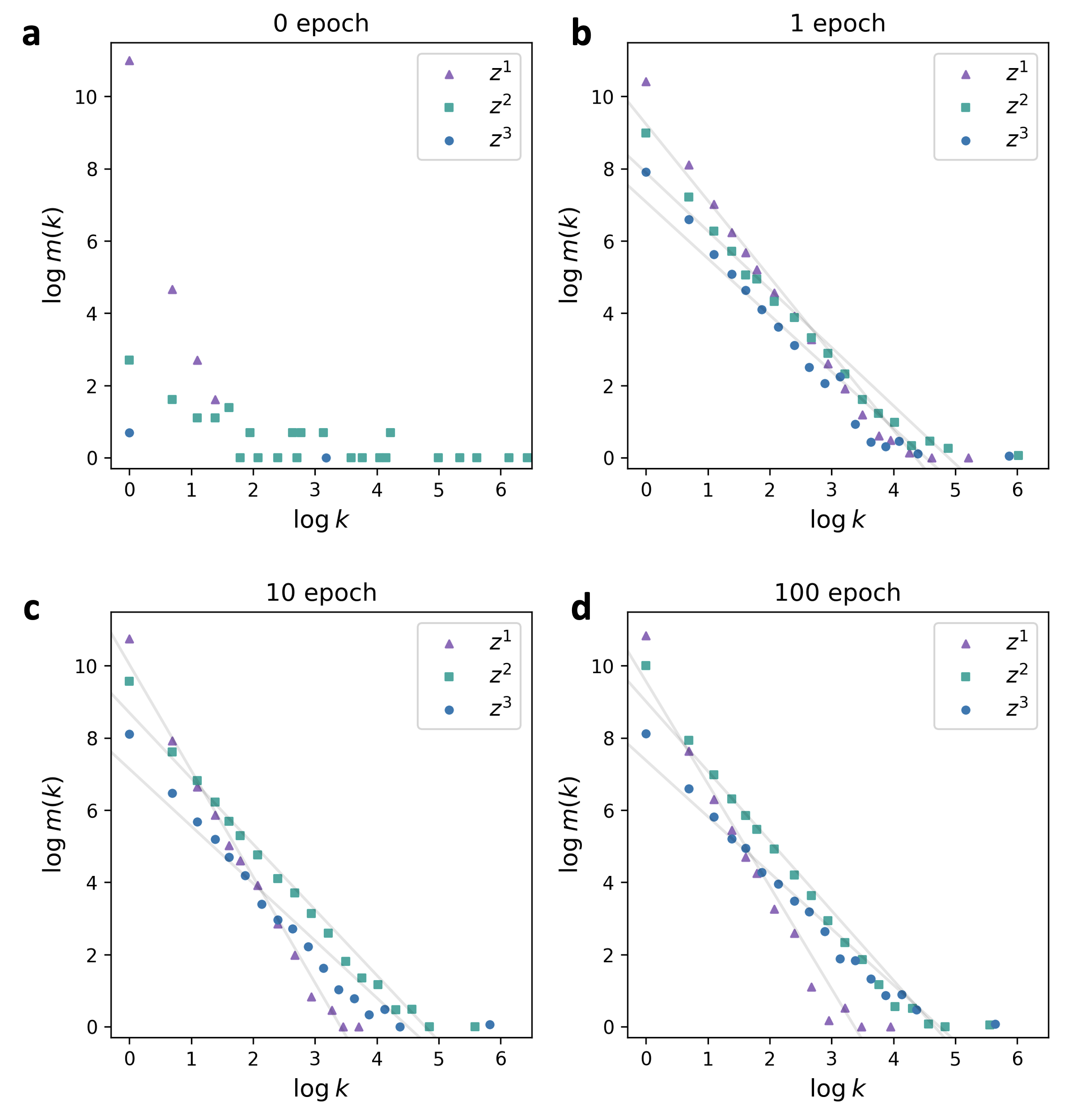}
\centering
\caption{\revi{(Color online) Data clustering during learning processes. Supervised learning of MNIST data through three internal representations of $z^1$, $z^2$, and $z^3$.
The frequency distributions of internal representations: $z^1$ (purple triangles), $z^2$ (green squares), and $z^3$ (blue circles) at (a) 0, (b) 1, (c) 10, and (d) 100 epochs.
}
\revf{Data beyond the window size, $\log k > 6.5$, were ignored.}
}
\label{fig4}
\end{figure}


\subsection{\revi{Learning and data}}
We also considered the question of whether these power laws originate from learning processes, or merely from data distributions. 

\revi{Suppose that neural networks start with random initial parameters before learning. 
Then, input data $x$ are transformed to random internal representations. Depending on parameter initialization schemes, a shallow internal representation $z^1$ sometimes displays a power-law scaling (Fig.~\ref{fig4}).
However, deep internal representation of $z^\mu$ for $\mu > 1$ experiences repeated transformations with random weights, which averages out the signal transfers, and converges into a few trivial representations.}
Therefore, the robust emergence of power laws clearly requires a learning process.
\revi{This excludes the possibility that the power laws may result from an artifact of random binarization of internal representations. However, it is surprising that a single epoch, which processes every training sample once, is sufficient to begin to show power-law scaling.}

\begin{figure}[t]
\includegraphics[scale=0.19]{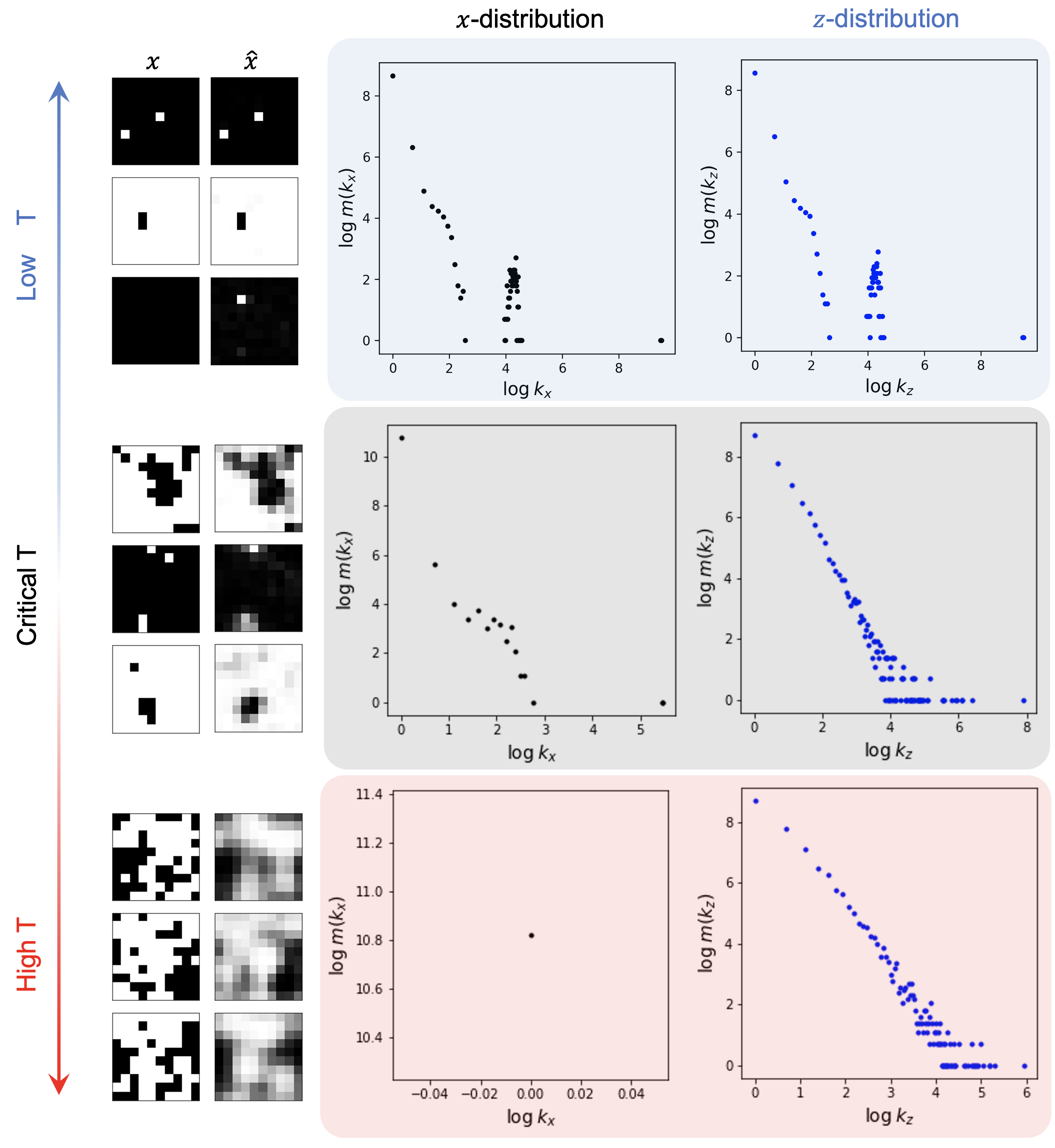}
\centering
\caption{(Color online) Self-supervised learning of Ising patterns. 
$10 \times 10$ lattice Ising patterns are used for input $x$ for a shallow autoencoder ($x\rightarrow z \rightarrow \hat{x}$).
Output $\hat{x}$ corresponds to reconstructed patterns.
A two-dimensional Ising model was used to generate 50,000 equilibrium samples at three temperatures (low $T = 1.53$, critical $T = 2.26$, high $T = 3.28$). 
Input $x$ and internal representation $z$ were characterized with different degeneracy distributions \revi{$m(k_x)$ and $m(k_z)$. For the binarization of $z$, we used a threshold of 0.4.}
}
\label{fig5}
\end{figure}

Next, the power laws should depend on original data distributions.
\revi{For example, when} data include identical samples, the distribution of corresponding internal representations is trivially influenced by the frequency of the identical samples.
Although the image data in our study did not include identical samples, one may speculate that they have complex structures which generate the power laws without a learning process.
\revi{To check this possibility, we examined structureless patterns $x$ sythesized by two-dimensional Ising models.}
The Ising model has equilibrium patterns of $x$ depending on its energy,  
\begin{equation}
    E(x) = - J \sum_{\langle i, j \rangle} x_i x_j,
\end{equation}
where $\langle i, j \rangle$ represents the nearest-neighboring pairs, and we set $J=1$.
Then, the realization probability of a pattern $x$ follows the Boltzmann distribution, as given below.
\begin{equation}
    p(x) = \frac{\exp[-E(x)/T]}{Z},  \phantom{MM} Z = \sum_x \exp[-E(x)/T].
\end{equation}
Depending on temperature $T$, diverse patterns can be generated.
Low temperature produces simple patterns including a few defects, whereas high temperature produces random patterns mixing black \revi{($x_i = 1$) and white ($x_i = -1$)} pixels.
At the critical temperature ($T \approx 2.26$) in the two-dimensional Ising model, complex patterns arose with long-range correlations between pixels.
\revf{For the simulation of the Ising model, We considered a $10 \times 10$ square lattice, and used a Monte-Carlo method with the Metropolis algorithm under low ($T=1.53$), critical ($T=2.26$), and high ($T=3.28$) temperatures.} 
We prepared 50,000 equilibrium samples at each temperature. Because low temperature produces simple patterns, many identical samples of $x$ were inevitably included in the low-temperature samples.
We input the diverse Ising patterns of $x$ to a shallow autoencoder  ($x\rightarrow z \rightarrow \hat{x}$) including a single hidden layer.
\revi{For the self-supervied learning of the Ising patterns, we changed $x_i = -1$ of white pixels into $x_i = 0$.}
The autoencoder was trained to reconstruct a pattern $\hat{x}$ identical with the input $x$. 
Given the input $x$ and transformed $z$, we obtained their frequencies $k_x$ and $k_z$, and then their degeneracy distributions of $m(k_x)$ and $m(k_z)$ for $x$ and $z$ (Fig. ~\ref{fig5}).
\revi{Note that we introduced new notations of $m(k_x)$ and $m(k_z)$ to distinguish the degeneracy $m(k)$ for $k_x$ and $k_z$.
As expected, the peculiar input distribution of $k_x$ due to identical samples at low temperature was trivially reflected in the distribution of $k_z$.
However, if sufficiently diverse patterns, generated above critical temperature, were used for learning, their internal representations follow power laws.
This result lends support to our observation of power laws not only for natural images, but also for synthetic structureless images, unless input data distributions exhibit unusual shapes with many identical samples.
}

\begin{figure}[t]
\includegraphics[scale=0.30]{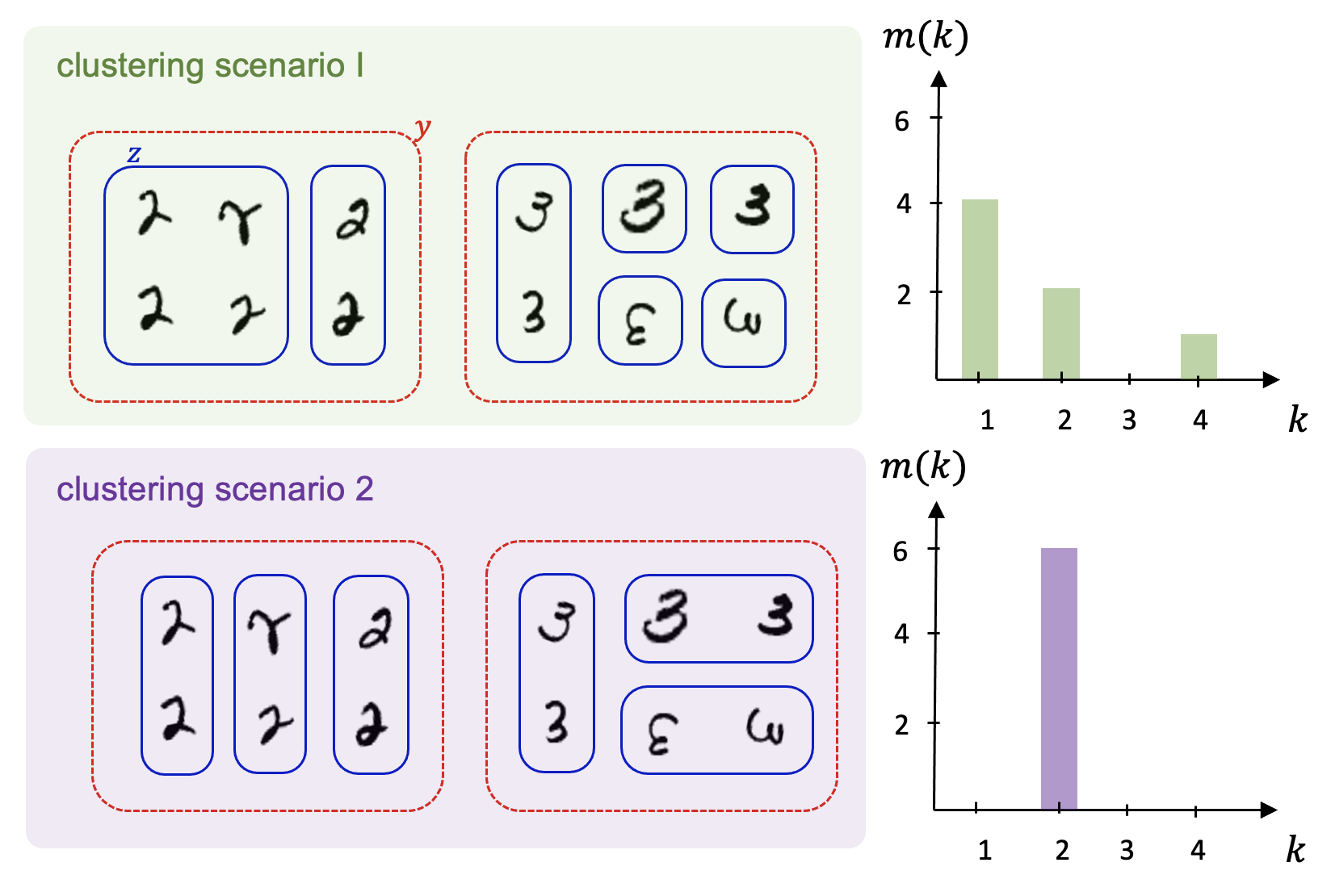}
\centering
\caption{(Color online) Cluster size distribution of machine learning. In supervised learning, an input image $x$ is represented by its compressed code $z$ (blue solid line), and then finally grouped into output $y$ (red dotted line). Cluster size distribution $m(k)$ of cluster size $k$ for two clustering scenarios. Two clustering scenarios give the same learning accuracy. Hence, we consider the question of which is more likely to occur.}
\label{fig6}
\end{figure}

\section{Theory}
\label{sec:theory}
The mechanism by which these power laws arise \revi{frequently} in various machine learning operating on diverse data poses an intriguing question. 
None of the learning algorithms were instructed as to the special shaping of $z$.
Our claim is that the power laws correspond to entropy-maximized distributions among possible distributions that satisfy given learning accuracies (Fig.~\ref{fig6}).
\revi{To progressively investigate this idea, we first review the information-theoretic concepts of {\it resolution} and {\it relevance}~\cite{marsili2013}, and the derivation of the scale-invariant hidden representations of RBMs~\cite{powerlaw}. Then, we extend this reasoning to explain the scale-invariant hidden representations for self- and authentic supervised learning, which is the major finding of this study.
}

\subsection{\revi{Resolution and relevance}}
\revi{Let us consider a random variable $z$, the frequency of which is $k_z$ with a total number of realizations $M=\sum_z k_z$.
The uncertainty of $z$ can be quantified using the Shannon entropy
\begin{equation}
\label{eq:HZ1}
    H(Z) = - \sum_z \frac{k_z}{M} \log \frac{k_z}{M}.
\end{equation}
Because $H(Z)$ quantifies the effective number of distinct realizations of $z$, it is referred to as {\it resolution}~\cite{marsili2013}.
In terms of coding theory, $H(Z)$ corresponds to a minimum description length for $z$~\cite{cubero2019}. 
Different realizations $z$ may exhibit the same frequency $k_z=k$. 
Unless we have any prior knowledge on $z$, the frequency $k_z$ may be the only feature extractable from $z$ at this point.
Then, it is natural to consider the degeneracy $m(k)$ of the frequency $k$. 
Given the degeneracy $m(k)$, we can reformulate Eq.~(\ref{eq:HZ1}) in terms of $k$-summation instead of $z$-summation as
\begin{equation}
\label{eq:HZ}
    H(Z) = -\sum_k \frac{k m(k)}{M} \log \frac{k}{M}
\end{equation}
with $M=\sum_k k m(k)$.
}

\revi{Now let us quantify the uncertainty of $k$ that a certain $z$ has a frequency $k_z = k$,
\begin{equation}
    H(K) = -\sum_k \frac{k m(k)}{M} \log \frac{km(k)}{M}.
\end{equation}
If one does not distinguish states $z$ that have the same frequency $k_z = k$, the variability of frequency $k$ can measure the amount of relevant information in data. Thus $H(K)$ is referred to as {\it relevance}~\cite{marsili2013}.
If every $z$ is distinct with the same frequency $k_z = 1$ and $m(1)=M$, we have no relevance $H(K)=0$, although we have a maximal resolution $H(Z) = \log M$. In contrast, if every $z$ is identical with $k_z = M$ and $m(M)=1$, we have also no relevance $H(K)=0$, but with zero resolution $H(Z) = 0$. Therefore, these two extreme cases correspond to a lack of frequency variability, in which we cannot distinguish $z$ in terms of their frequency $k_z$.
}

\subsection{\revi{Unsupervised learning}}
\revi{The information measures of resolution and relevance have been adopted to explain the scale-invariant hidden representations of RBMs~\cite{powerlaw, cubero2019}.
The graphical model of RBMs consists of visible and hidden units of $x$ and $z$. RBMs are trained to have a large model probability $p(x,z)$ with a good pairing of data $x$ and hidden representation $z$~\cite{hinton2006}.
We can interpret $z$ as emergent labels for $x$.
This allows us to define a group of $x$, which have the same label $z$, as a cluster.
In particular, under compressing conditions when the dimension of $z$ is smaller than the dimension of $x$, similar $x$ are grouped together with label $z$.
Then, $k_z$ denotes the size of the $z$-labeled cluster, and $m(k)$ corresponds to the distribution of cluster sizes.
}

What is an expected distribution of the cluster sizes?
RBMs do not impose any constraint on the shaping of the size distribution during the learning process.
Song {\it et al.} found that the size distribution $m(k)$ follows power laws, and derived that the power laws correspond to the most likely distribution at a fixed resolution of $z$~\cite{powerlaw}.
The scale-invariant distribution maximizes the uncertainty of the size $k$ of a cluster to which a specific $x$ belongs.
This constrained optimization can be formulated using Lagrange multipliers, as given below.
\begin{equation}
    \label{eq:unsupervised}
    \mathcal{L} = H(K) + \beta \bigg(H(Z) - R \bigg) + \alpha \bigg(\sum_k \frac{k m(k)}{M} - 1 \bigg).
\end{equation}
The Lagrange multiplier $\alpha$ controls the normalization for $k$-distribution, while $\beta$ controls the resolution \revf{of hidden representation $z$ for a fixed value of $H(Z)=R$.}
The maximum frequency variability $H(K)$ subject to the two constraints is obtained at the variation condition of $\delta \mathcal{L}/\delta m(k) = 0$. 
The optimal condition leads to the power-law size distribution
\begin{equation}
\label{eq:powerlaw}
    m(k) \propto k^{-\beta-1}.
\end{equation}
Moreover, the scale-invariant hidden representation $z$ \revf{is good enough to make the model probability $p(x)= \sum_z p(x, z)$ close to the data probability $\hat{p}(x)$.}

\subsection{\revi{Supervised learning}}
\revi{The goal of supervised learning is faithful reproduction of true labels $y$ given input $x$.
For multi-layer neural networks, $x$ is transformed to hidden representations $z$, and then, $z$ is again transformed to output $\hat{y}$. The supervised learning optimizes an appropriate representation $z$ to produce $\hat{y}$, which is ultimately used to predict true $y$.
The learning accuracy can be estimated by the mutual information between internal representation $z$ and true label $y$,
\begin{align}
    I(Z;Y)=H(Y)-H(Y|Z),
\end{align} 
that quantifies how much uncertainty $H(Y)$ of $y$ is reduced by knowing $z$. The information gain corresponds to learning accuracy of the internal representation $z$.
Here, we focus on the neural networks, the internal layers of which have smaller sizes with increasing distance from the input layer.
The compressing condition provides a coarse-grained representation $z$ for $x$.
In particular, if we discretize $z$, we can again interpret a group of $x$ with the same hidden state $z$, as a cluster. 
Now, we derive the most likely distribution of hidden representation $z$ that guarantees a certain learning accuracy as $I(Z;Y) = R'$.
Like the objective for unsupervised learning in Eq.~(\ref{eq:unsupervised}), the objective for supervised learning can be formulated as
\begin{equation}
    \label{eq:supervised}
    \mathcal{L'} = H(K) + \beta \bigg(I(Z;Y) - R' \bigg) + \alpha \bigg(\sum_k \frac{k m(k)}{M} - 1 \bigg).
\end{equation}
Among possible representations $z$ satisfying a given learning accuracy, we find a representation $z$ that maximizes the uncertainty of the cluster size.
Caution is required to understand that this exploration is different from the optimization of learning algorithms.
We seek the most flexible representation $z$ providing the largest frequency variability $H(K)$, subject to a fixed learning accuracy.
Thus, we explore snapshots of $z$ at any learning status.
Indeed, we observed that supervised learning exhibited power-law distributions of $m(k)$ at any given learning accuracy during learning epochs (Fig.~\ref{fig4}).
}

\subsubsection{\revi{Self-supervised learning}}
\revi{We first consider the self-supervised learning of autoencoders in which labels are input themselves as $y=x$.
The mutual information of autoencoders is simply $I(Z;X)=H(Z)-H(Z|X)=H(Z)$, where the conditional entropy $H(Z|X)=0$ vanishes because the hidden representation is a deterministic function $z(x)$ of $x$ in autoencoders.
This condition makes Eq.~(\ref{eq:supervised}) identical to Eq.~(\ref{eq:unsupervised}) with $\mathcal{L}'=\mathcal{L}$ and $R'=R$.
We have already established that the optimal size distribution follows power laws in Eq.~(\ref{eq:powerlaw}) that maximizes $\mathcal{L}$.
This explains why power laws naturally arise in self-supervised learning.
}

As mentioned earlier, the unfolded structure of RBMs can be interpreted as constrained autoencoders~\cite{AE_RBM}. The forward and backward propagation of $x \leftrightarrow z$ corresponds to the information flow of $x \rightarrow z \rightarrow \hat{x}$.
\revf{To be specific, we consider a vector of $x = (x_1, x_2, \cdots)$ and a vector of $z = (z_1, z_2, \cdots)$.
The autoencoders have a constraint that the encoder weight parameter $w_{ij}$ for $x_i \rightarrow z_j$ remains the same with the transpose of the decoder weight parameter $w_{ji}$ of $z_j \rightarrow \hat{x}_i$.}
Then, the scale-invariant hidden representation of RBMs can be understood either as (i) maximizing $H(K)$ subject to a certain resolution $H(Z)=R$, or (ii) maximizing $H(K)$ subject to a certain learning accuracy $I(Z;X)=R'$.

\subsubsection{\revi{Authentic supervised learning}}
\revi{Next, we show that the previous conclusion on self-supervised learning is also applicable to authentic supervised learning.
The mutual information can be decomposed as
\begin{equation}
\label{eq:MI}
    I(Z;Y) = H(Y) + H(Z) - H(Y,Z).
\end{equation}
The first term $H(Y)$ is constant, because labels $y$ are given as data in supervised learning.
As a result, $H(Y)$ is trivially independent of $z$ and $m(k)$.
The second term $H(Z)$ depends on $m(k)$ as shown in Eq.~(\ref{eq:HZ}).
The third term is the entropy for the joint frequency $k_{y,z}$,
\begin{equation}
    H(Y,Z) = - \sum_{y,z} \frac{k_{y,z}}{M} \log \frac{k_{y,z}}{M}.
\end{equation}
Now, we derive that $H(Y,Z)$ does not explicitly depend on $m(k)$.
At first, the independence of $H(Y,Z)$ from $m(k)$ appears non-trivial, as both $H(Y,Z)$ and $m(k)$ depend on $z$.
For the self-supervised learning ($Y=X$), we have a constant $H(X, Z) = \log M$, if every $x$ is distinguished ($k_{x,z} = 1$).
It is clear that $H(X, Z)$ is always constant independently of $m(k)$.
}

\begin{figure}[t]
\includegraphics[scale=0.3]{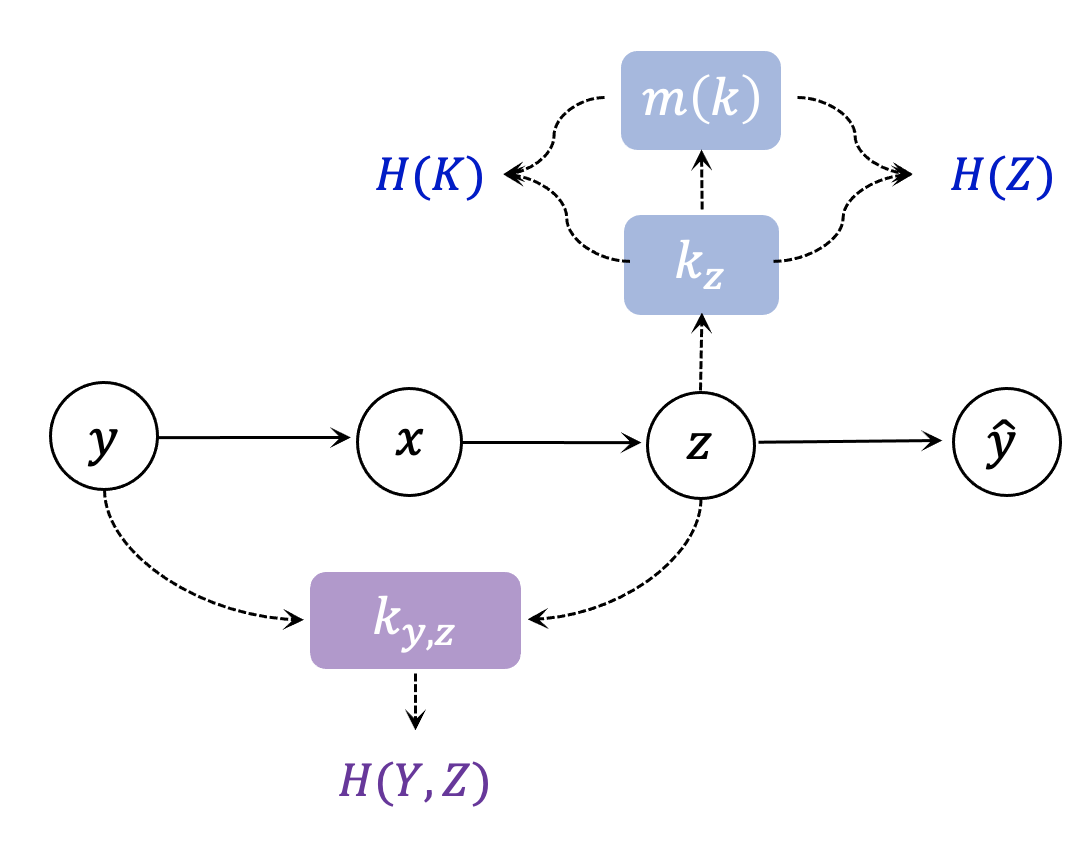}
\centering
\caption{\revi{(Color online) Causal graph of supervised learning.}
Solid arrows represent explicit transformation from parent to child nodes, whereas dotted arrows represent implicit transformation from a set of parent variables to child nodes.
}
\label{fig7}
\end{figure}

\begin{figure}[t]
\includegraphics[scale=0.3]{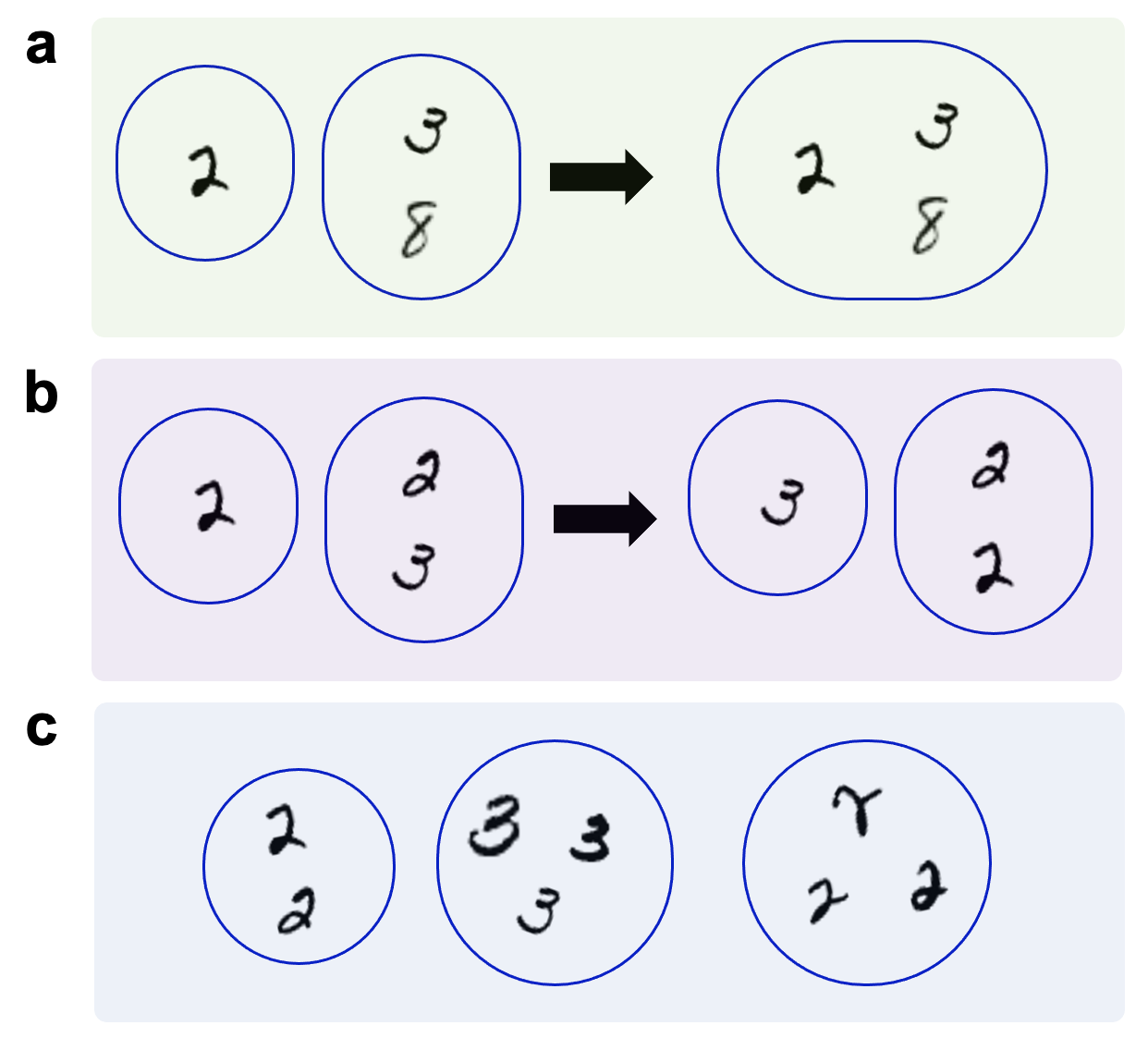}
\centering
\caption{\revi{(Color online) Schematic diagrams of data clustering with labels and internal representations. MNIST digit images $x$ have true labels (2, 3, and 8), and they are grouped with their internal representations $z$ (blue circles). 
(a) A scenario in which $k_z$ changes from $k_z=\{1, 2\}$ to $k_z=\{3\}$, but $k_{y, z} =1$ does not change. Therefore, the size distribution $m(k)$ changes, but $H(Y,Z)$ does not change.
(b) A scenario in which $k_z = \{1, 2\}$ does not change, but $k_{y, z}$ changes from $k_{y, z} = 1$ to $k_{y, z} = \{1, 2\}$. Therefore, the size distribution $m(k)$ does not change, but $H(Y, Z)$ changes.
(c) A pure clustering scenario in which every $x$ for a given $z$ has the same label $y$.
}
}
\label{fig8}
\end{figure}

\revi{For the authentic supervised learning, let us imagine a causal graph between variables (Fig.~\ref{fig7}).
The number of realizations of hidden representation $z$ is the frequency $k_z$, the degeneracy of which is $m(k)$. The number of realizations for $y$ and $z$ is $k_{y, z}$, which is used to compute $H(Y,Z)$. Therefore, $z$ is a confounder that affects both $m(k)$ and $H(Y,Z)$ in the causal graph. 
In terms of causality~\cite{pearl2009}, $H(Y,Z)$ and $m(k)$ are called {\it d}-separate for fixed $z$,
\begin{equation}
    H(Y,Z) \phantom{l} \indep \phantom{l} m(k) \phantom{l}| \phantom{l} z,
\end{equation}
which means that $m(k)$ does not explicitly affects $H(Y,Z)$ and vice versa.
The functional independence can be further demonstrated in two situations. The first situation is that $m(k)$ changes, but $H(Y,Z)$ does not change (Fig.~\ref{fig8}a).
The second situation is that $H(Y,Z)$ changes, but $m(k)$ does not change (Fig.~\ref{fig8}b).
}

\revi{
Thus far, we have confirmed that $H(Y)$ and $H(Y,Z)$ in the learning accuracy of Eq.~(\ref{eq:MI}) do not explicitly depend on $m(k)$. This results in $\delta \mathcal{L}'/\delta m(k) = \delta \mathcal{L}/\delta m(k)$. 
This explains why authentic supervised learning exhibits a power-law distributions of $m(k)$ as in self-supervised learning.
}

\revi{It is noteworthy to mention one exceptional situation, in which every cluster has data $x$ with pure labels $y$ (Fig.~\ref{fig8}c). This situation is relatively unlikely to occur in practice in machine learning with large noisy data. The ideal clustering with no impurity leads to $H(Y, Z) = H(Z) + H(Y|Z) =H(Z)$, because labels do not include additional information beyond the hidden representation $z$. 
Substituting this result into Eq.(\ref{eq:MI}), we obtain $I(Z;Y)=H(Y)$. This makes $H(K)$ the only $m(k)$-dependent term of $\mathcal{L}'$ in Eq.~(\ref{eq:supervised}). Then, $m(k) \propto k^{-1}$ maximizes $\mathcal{L}'$. This solution corresponds to the power-law distribution with $\beta = 0$. The zero value of Lagrange multiplier nullifies the constraint for the learning accuracy in $\mathcal{L}'$.
}

\revi{In summary, except for the pure clustering, the objective $\mathcal{L}'$ for supervised learning has only two $m(k)$-dependent terms of $H(K)$ and $H(Z)$. This conclusion is the same with $\mathcal{L}$ for unsupervised learning.
The objective functional, $\mathcal{L} = H(K) + \beta H(Z)$, has been intensively studied by Marsili {\it et al.}~\cite{marsili2013,haimovici2015,powerlaw,cubero2019,duranthon2021, marsili2022}. 
Assuming that data follows a Boltzmann distribution, $k/M \propto \exp(-E(k))$, as in equilibrium thermodynamics, the entropy of $Z$ can be interpreted as an internal energy by ignoring a constant shift.
\begin{equation}
    H(Z) = -\sum_k p(k) \log \frac{k}{M}= \sum_k p(k) E(k) = U,  
\end{equation}
where $p(k) = km(k)/M$.
Then, the Shannon entropy $H(K)$ can be interpreted as thermodynamic entropy.
\begin{equation}
    H(K) = -\sum_k p(k) \log p(k) = S.
\end{equation}
Interpreting $\beta = -1/T$ as negative inverse temperature, one can define thermodynamic free energy.
\begin{equation}
    \mathcal{F} = -T\mathcal{L} = U - TS.
\end{equation}
Then, the maximization of $\mathcal{L}$ corresponds to the minimization of free energy $\mathcal{F}$.
The optimal distribution $m(k)$, satisfying $\delta \mathcal{F} / \delta m(k) =0$, has thus been derived to follow the scale-invariant power laws, $m(k) \propto k^{-\beta - 1}$~\cite{marsili2013}.
This explains why power laws naturally arise in both unsupervised and supervised learning.
This result could thus be referred to as an analogue of the second law of thermodynamics in machine learning.
}

\section{Summary}
\label{sec:summary}
\revi{We have studied the internal representations $z$ of data in machine learning. 
Song {\it et al.} first observed that restricted Boltzmann machines have special representations with a few frequent $z$ and many rare $z$, the frequency distributions of which follow power laws~\cite{powerlaw}.  
In this study, we showed that the scale-invariant representations are observed not only in unsupervised learning, but also in supervised learning.
Furthermore, we have derived that the scale invariance can naturally arise in machine learning using information theory.
The critical representations correspond to entropy-maximized encodings given learning accuracies.
If we define a group of data $x$ that have the same $z$ (compressed codes or emergent labels), the frequencies of $z$ can be interpreted as cluster sizes of $x$.
Then, the maximum uncertainty of the cluster size distribution implies that the size of a cluster, to which a certain data $x$ belongs, can be most flexibly determined.
Therefore, at any given learning accuracy, $z$ can show the criticality.
}

In this study, we have examined compressing structures of neural networks with classical architectures such as multi-layer perceptrons, vanilla autoencoders, convolutional neural networks, and restricted Boltzmann machines, although recent deep learning considers infinitely-wide networks and overparameterized models as well~\cite{lee2017, oymak2019, geiger2020}.
It remains a topic of future research to explore the application of our conjecture of {\it the second law of machine learning thermodynamics} in modern architectures
\revf{of large-scale generative models with a compressed latent space.}

The power laws, also known as the Pareto principle, have been ubiquitously observed in social and biological data including real neural activities~\cite{newman2005,hidalgo2014,tkavcik2015,james2018,tomen2019}. Unlike the symbolic property of criticality in statistical mechanics, the emergence of criticality in those data does not require fine tuning~\cite{schwab2014, aitchison2016, touboul2017}. Schwab et al. have shown that multivariate systems can generate such criticality without fine tuning when latent variables are involved in the systems~\cite{schwab2014}.
We note that whereas these studies focused on the statistics of observed data $x$ in real neural networks, the present study has explained the statistics of internal representation $z$ in artificial neural networks.
The criticality of existing data that comprise the natural world and the emergence of criticality during the learning process, discussed in this study, are of considerable intellectual interest. A further investigation of criticality may be expected to stimulate cross-fertilization between the fields~\cite{savage2019}.

\section*{Acknowledgment}
We thank Sang-Gyun Youn, Vipul Periwal, Deok-Sun Lee, Changbong Hyeon, and Matteo Marsili for helpful discussions and comments.
This work was supported in part by the National Research Foundation of Korea (NRF) grant (Grant No. 2021R1A2C2012350) (S.L.), the New Faculty Startup Fund from Seoul National University, and the NRF grant funded by the Korea government (MSIT) (Grant No. 2019R1F1A1052916) (J.J.).

\bibliographystyle{apsrev4-1}
\bibliography{reference}

\end{document}